\pdfoutput=1

\documentclass[11pt]{article}

\usepackage[final]{acl}
\usepackage{adjustbox}
\usepackage{enumitem}
\usepackage{tabularx}
\usepackage{booktabs}
\usepackage{multicol}
\usepackage{wasysym}
\usepackage[multiple]{footmisc}
\usepackage{makecell}
\usepackage{tikz}
\usepackage[utf8]{inputenc}
\usepackage{textcomp}
\usepackage{xcolor}
\usetikzlibrary{arrows.meta, positioning, shapes.geometric}
\usepackage{CJKutf8}
\usepackage[toc,page]{appendix}
\usepackage{times}
\usepackage{latexsym}
\usepackage{multirow}

\usepackage{array}
\usepackage{booktabs}
\usepackage{geometry}
\usepackage{caption}

\usepackage{microtype}

\usepackage{inconsolata}

\usepackage{graphicx}
\usepackage{blindtext}

\title{PolyNorm: Few-Shot LLM-Based Text Normalization for Text-to-Speech}

\author{
Michel Wong\thanks{Correspondence: \texttt{michel.wong@apple.com}}, 
Ali Alshehri, 
Sophia Kao, 
Haotian He \\
\texttt{\{michel.wong, a\_alshehri, sophia\_kao, haotian\}@apple.com}
\\\\Apple}

\begin{document}
\maketitle
\begin{abstract} 
Text Normalization (TN) is a key preprocessing step in Text-to-Speech (TTS) systems, converting written forms into their canonical spoken equivalents. Traditional TN systems can exhibit high accuracy, but involve substantial engineering effort, are difficult to scale, and pose challenges to language coverage, particularly in low-resource settings. We propose PolyNorm, a prompt-based approach to TN using Large Language Models (LLMs), aiming to reduce the reliance on manually crafted rules and enable broader linguistic applicability with minimal human intervention. Additionally, we present a language-agnostic pipeline for automatic data curation and evaluation, designed to facilitate scalable experimentation across diverse languages. Experiments across eight languages show consistent reductions in the word error rate (WER) compared to a production-grade-based system. To support further research, we release PolyNorm-Benchmark, a multilingual data set covering a diverse range of text normalization phenomena.
\end{abstract}

\section{Introduction}

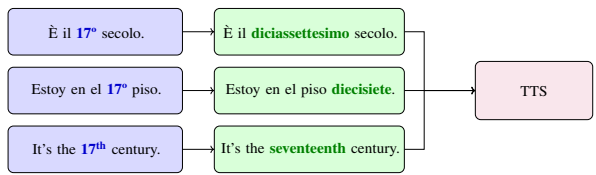
\begin{figure}[ht]
\centering
\begin{adjustbox}{width=\columnwidth, center}
\begin{tikzpicture}[node distance=0.3cm and 0.8cm, font=\normalsize]
  \node[draw, fill=blue!15, rounded corners, text width=4.2cm, minimum height=1.2cm, align=center] (input1) {È il \textcolor{blue!80!black}{\textbf{17º}} secolo.};
  \node[draw, fill=blue!15, rounded corners, below=of input1, text width=4.2cm, minimum height=1.2cm, align=center] (input2) {Estoy en el \textcolor{blue!80!black}{\textbf{17º}} piso.};
  \node[draw, fill=blue!15, rounded corners, below=of input2, text width=4.2cm, minimum height=1.2cm, align=center] (input3) {It's the \textcolor{blue!80!black}{\textbf{17\textsuperscript{th}}} century.};
  \node[draw, fill=green!15, rounded corners, text width=4.6cm, minimum height=1.2cm, right=of input1, align=center] (norm1) {È il \textcolor{green!50!black}{\textbf{diciassettesimo}} secolo.};
  \node[draw, fill=green!15, rounded corners, text width=4.6cm, minimum height=1.2cm, right=of input2, align=center] (norm2) {Estoy en el piso \textcolor{green!50!black}{\textbf{diecisiete}}.};
  \node[draw, fill=green!15, rounded corners, text width=4.6cm, minimum height=1.2cm, right=of input3, align=center] (norm3) {It's the \textcolor{green!50!black}{\textbf{seventeenth}} century.};
  \node[draw, fill=purple!10, rounded corners, minimum width=3cm, minimum height=1.5cm, right=1.8cm of norm2, align=center] (tts) {TTS};
  \draw[->] (input1.east) -- ++(0.5,0) -- (norm1.west);
  \draw[->] (input2.east) -- ++(0.5,0) -- (norm2.west);
  \draw[->] (input3.east) -- ++(0.5,0) -- (norm3.west);
  \draw[->] (norm1.east) -- ++(0.5,0) |- (tts.west);
  \draw[->] (norm2.east) -- ++(0.5,0) -- (tts.west);
  \draw[->] (norm3.east) -- ++(0.5,0) |- (tts.west);

\end{tikzpicture}
\end{adjustbox}
\caption{The ambiguous token 17° is normalized differently based on context and language.}
\label{fig:text_normalization}
\end{figure}

TN transforms written input, often dense with numbers, abbreviations, and special characters, into fluent speech-friendly text. Early TN systems relied primarily on rule-based approaches using weighted finite-state transducers (WFSTs) \cite{sproat2001normalization}. Although such systems have historically achieved high accuracy, they depend heavily on manual rules and extensive human verification, making them expensive and time-consuming to develop and maintain. These challenges are even more pronounced when dealing with low-resource or morphologically rich languages, such as Arabic and Polish \cite{mosquera2012towards,poswiata2019numbers}, where linguistic diversity complicates the creation of rules. Beyond the scarcity of resources, TN systems must also handle contextual ambiguities that arise within or across languages. The same surface form can be normalized differently depending on both the linguistic context and the target language, as illustrated in Figure~\ref{fig:text_normalization}\footnotemark.

\footnotetext{
“È il 17° secolo.” translates to “It’s the 17\textsuperscript{th} century.”
}

\footnotetext{
“Estoy en el 17º piso.” translates to “I am on the 17\textsuperscript{th} floor.”
}

Later development lead to TN systems that are based on statistical and hybrid data-driven approaches \cite{gralinski2006text,zhu2007unified,sproat2016rnn}. For example, \citet{gralinski2006text} treated TN as a machine translation problem, mapping textual representations into their spoken equivalents, while \citet{zhu2007unified} introduced a unified tagging approach using Conditional Random Fields (CRF) to simultaneously handle multiple TN subtasks, emphasizing their interdependencies. Recently, large language models (LLMs) have shown great potential in addressing these limitations by leveraging their broad linguistic knowledge. \citet{zhang2024chatboringproblemsstudying} used GPT models to perform text normalization in few-shot scenarios, incorporating self-consistency reasoning and linguistically informed prompt engineering, achieving error rates approximately 40\% lower than those of production-level WFST systems.

Building on prior research that focused on English and GPT models, this work introduces PolyNorm, an LLM-assisted TN framework that leverages prompting and contextual learning, and compares multiple LLMs across diverse languages. Our research pursues two primary goals: (1) developing a sustainable, cost-effective TN solution with minimal human involvement, and (2) creating a reliable, language-agnostic process for automatic data curation and evaluation applicable across both high- and low-resource languages, offering a scalable and accessible TN solution for modern TTS systems.

\section{Data}
Our research focuses on American English, German, French, Mexican Spanish, Italian, Lithuanian, Japanese, and Mandarin Chinese. 

We start with a subset of the English Kestrel dataset~\cite{sproat2015kestrel} comprising 14 Kestrel categories extracted from the first file of the dataset, which contains approximately 880{,}000 lines. For English, to construct a balanced subset, we select a total of 1,400 test cases, removing all sentences that lack normalization targets and retaining the first 100 examples per category from the Kestrel dataset for benchmarking purposes. We initially experimented with translating both the original and normalized English texts into the target languages with NLLB-200 model~\cite{nllb2022}. However, we found most translated texts unsuitable as ground truth due to inconsistent quality, translation and inadequate localization of the normalized forms. In addition, the limited categories do not fully cover the scope of normalization phenomena that we aim to evaluate. Therefore, instead of using the entire subset for benchmarking, we reserved some of the higher-quality data, reviewed and verified by language experts, as part of the prompt examples, which will be discussed in the Methodology section.

To address these challenges, we use DeepSeek-R1~\cite{guo2025deepseek} to generate datasets comprising unnormalized and normalized text pairs as our new multilingual benchmark suite, \emph{PolyNorm-Benchmark}\footnote{https://github.com/apple/ml-speech-polynorm-bench}, designed to ensure consistency across languages and taxonomies. Each dataset spans 27 normalization categories (listed in Table~\ref{tab:tn_categories}) built on Kestrel categories and expanded with new classes, comprising 20 examples per category for a total of 540 high-quality data points per language. For Chinese and Japanese, we define abbreviations as truncated forms or initial-letter substitutions, since conventional definitions may not fully capture their use in conversational language.

\begin{table}[t]
\centering
\small
\begin{tabular}{@{}p{0.45\columnwidth}@{} p{0.45\columnwidth}@{}}
\toprule
\multicolumn{2}{c}{\textbf{List of Text Normalization Categories}} \\
\midrule
Cardinal & Date \\
Decimal & Ordinal \\
Fraction & Time \\
Currency & Unit (Measure) \\
Address & Acronym/Initialism \\
ISBN & Biological Classification \\
Roman Numeral & Telephone \\
Sports Score & Mathematical Expression \\
Symbol & Abbreviations\\
Chemical Formula & Legal Reference \\
Vehicle/Product Code & Geographic Coordinates \\
Version Number & License/Serial Number \\
Musical Notation & Stock Ticker \\
Electronic (URL/Email) & \\
\bottomrule
\end{tabular}
\caption{List of 27 text normalization categories used for multilingual data construction.}
\label{tab:tn_categories}
\end{table}

The initial outputs of DeepSeek serve as a foundation, then each example is subsequently edited and verified by internal language experts to ensure the highest standards of linguistic precision, naturalness, and overall quality. The orthography and formatting of synthetic data follow the conventions of target languages. For example, in French, decimal separator in currency is written with a comma instead of a period, with the currency symbol after the number:

\small
\begin{quote}
\textbf{Orig:} \texttt{Le taux de change est de 1€ = 1,10\$.} \\
\textbf{Norm:} \texttt{Le taux de change est de un euro égale un virgule dix dollars.}
\end{quote}

\begin{quote}
\textbf{Orig:} \texttt{El 1º de Enero.} \\
\textbf{Norm:} \texttt{El primero de Enero.}
\end{quote}

\begin{quote}
\begin{CJK*}{UTF8}{gbsn}
\textbf{Orig:} \texttt{容量は500mL。} \\
\textbf{Norm:} \texttt{ヨウリョウはゴヒャクミリリットル。}
\end{CJK*}
\end{quote}
\normalsize

\section{Methodology}

We define a normalization error as any output in which the normalized text deviates from the expected conventional spoken form. For example, rendering “Dr” as “Doctor” instead of “Drive” in an address, or normalizing “12:30” as “twelve thirty AM” when the time format is ambiguous.

For clarity and consistency, we standardize normalization to common conventions across languages when applicable. For instance, in American English, we follow the month-day-year format, consistent with our TTS production system expectations. Years like "2020" are normalized as \texttt{twenty twenty}, and dates such as "4/18" are normalized as \texttt{april eighteenth}.

\small
\begin{quote}
\textbf{Orig:} \texttt{05/20/2023} \\
\textbf{Norm (accepted):} \texttt{may twentieth twenty twenty three} \\
\textbf{Norm (rejected):} \texttt{the twentieth of may two thousand and twenty three}
\end{quote}
\normalsize

Certain differences are not considered errors in our evaluation, as they do not affect the semantic or phonetic realization of the text. These include casing, optional spacing (e.g., in German compound words), and orthographic variants such as "ß" vs. "ss" in specific German contexts, ensuring the evaluation focuses on differences affecting meaning, pronunciation, or intelligibility. To identify such issues and iteratively refine our system, we collaborate with language experts who review LLM outputs on our development sets and flag inaccuracies or inconsistencies, which inform benchmark corrections, in-context learning (ICL) examples refinements, and system iterations.

We adopt a few-shot prompting strategy using ICL examples to guide the LLM in performing text normalization across all languages. This approach allows the model to generalize from a small number of curated examples across various categories, from numerical expressions to acronyms, eliminating extensive fine-tuning or rule-based engineering.

Our prompt design follows a structured, step-by-step and language-agnostic format composed of three components: 1) \textbf{Instruction Prompt}, which defines the normalization task, 2) \textbf{In-Context Learning Examples}, which demonstrate how different types of text should be normalized, and 3) \textbf{Target Unnormalized Input}, to which the model applies learned patterns. PolyNorm uses a standardized English instruction prompt with localized ICL examples tailored to each language’s linguistic and stylistic norms.

\subsection{ICL Examples}

Few-shot prompting with ICL offers a lightweight, scalable solution for multilingual text normalization, relying on high-quality, language-specific examples. The effectiveness of this approach depends on the quality and relevance of the ICL examples provided to the model. We use a unified prompt format across eight languages, varying only the localized examples—except for Japanese, where a supplementary prompt guides katakana output. Our curated ICL set, excluded from PolyNorm-Bench, includes 80-100 high-quality machine-translated Kestrel data and DeepSeek synthetic examples (e.g., favoring “twenty nineteen” over “two thousand nineteen” for "2019" in English), each normalized and validated by experts to ensure stylistic consistency across domains.

\subsection{Error Analysis and Hillclimbing Iteration}

We analyzed discrepancies between expert-verified development sets and LLM outputs from GPT models to identify systematic weaknesses. Common errors included incorrect numeral expansions and inconsistent handling of language-specific formats such as date and currency. To address these issues, we identified categories or patterns where the model underperformed, then revised or added ICL examples focused on these error types, improving the model’s output with each iteration. This feedback loop refined normalization via prompt tuning and better examples, reducing errors and improving cross-lingual consistency.

\section{Results}

We evaluated GPT-4o-mini and GPT-4o as candidate systems against Apple Siri production rule-based normalization baseline system. Because it is proprietary, we cannot release code or provide external access. However, to help readers judge strength and fairness, we report baseline scores for a representative subset of seven categories across all eight languages in Table~\ref{tab:baseline:results} below\footnote{Table~\ref{tab:baseline:tn-errors} in Appendix~\ref{app:baseline:examples} shows side-by-side examples illustrating some incorrect baseline outputs}. It is important to note that the baseline predates our benchmark and was not tuned to it.

All systems were tested on PolyNorm across eight languages listed in Table~\ref{tab:results}. We report the overall performance with Word Error Rate (WER) and BLEU, which reflects how closely the model’s output matches the expected spoken form and relative surface similarity to the reference respectively. For non-whitespace languages (Chinese and Japanese), we report Character Error Rate instead of WER to avoid conflating tokenization errors with true normalization errors. PolyNorm demonstrates substantial advantages in terms of flexibility, coverage, and the ability to handle previously intractable normalization challenges with minimal manual human intervention. Results in Table~\ref{tab:results} show improvements in both WER and BLEU across the two LLM systems compared to the rule-based baseline. GPT-4o achieves a WER ranging from 4.17\% to 7.88\% for all languages.

\begin{table*}[t]
\centering
\small
\begin{tabularx}{\textwidth}{l *{6}{>{\centering\arraybackslash}X}}
\toprule
\textbf{Language} & \multicolumn{2}{c}{\textbf{Baseline}} & \multicolumn{2}{c}{\textbf{GPT-4o-mini}} & \multicolumn{2}{c}{\textbf{GPT-4o}} \\
\cmidrule(lr){2-3} \cmidrule(lr){4-5} \cmidrule(lr){6-7}
 & \textbf{WER (\%)} & \textbf{BLEU (\%)} & \textbf{WER (\%)} & \textbf{BLEU (\%)} & \textbf{WER (\%)} & \textbf{BLEU (\%)} \\
\midrule
German            & 10.74 & 60.83 & 7.18 & 78.35 & \textbf{4.17} & \textbf{84.92} \\
American English  & 9.84 & 70.84 & 6.60 & 85.02 & \textbf{4.28} & \textbf{89.85} \\
Mexican Spanish   & 11.92 & 55.74 & 11.19 & 62.49 & \textbf{7.69} & \textbf{71.90} \\
French            & 9.72 & 69.02 & 7.70 & 79.84 & \textbf{5.65} & \textbf{86.18} \\
Italian           & 15.02 & 55.14 & 8.33 & 72.81 & \textbf{4.56} & \textbf{84.96} \\
Lithuanian        & 10.04 & 67.12 & 10.54 & 64.36 & \textbf{6.99} & \textbf{73.96} \\
Mandarin Chinese  & 11.36 & 69.63 & 6.65 & 79.04 & \textbf{5.05} & \textbf{83.11} \\
Japanese          & 17.49 & 47.76 & 10.78 & 68.32 & \textbf{7.88} & \textbf{77.89} \\
\bottomrule
\end{tabularx}
\caption{WER and BLEU scores across 8 languages using PolyNorm-Benchmark. Best scores per row are in bold.}
\label{tab:results}
\end{table*}

We observe that domain-specific constructs, such as website URLs and usernames, which rely on static rules, often pose challenges for tokenization and rule-based normalization due to their inherent unpredictability and rapid evolution. PolyNorm addresses these challenges with the contextual capabilities of LLMs to intelligently segment such expressions. For instance, \texttt{https://www.mediacityuk.co.uk} is normalized as \texttt{h t t p s colon slash slash media city u k dot co dot u k}.

Another observation of PolyNorm’s context awareness over rule-based systems is its handling of dashes between numbers. In sports scores, a dash is typically verbalized as "to" (e.g., "3-2" as "three to two"), whereas in phone numbers it is ignored and the digits are read individually. A rule-based system would require an explicit and often verbose set of context-specific rules to make this distinction, while PolyNorm can infer the correct interpretation directly from context, even in novel or ambiguous formats.
\section{Discussion}

\subsection{Impact}
Training a reliable TTS system can require millions of high-quality data points. While effective in controlled environments, traditional TN pipelines, heavily rule-based, demand ongoing annotation and patching, often taking months per language and struggling to generalize or scale to noisy, ambiguous, or evolving inputs. PolyNorm replaces this manual loop with an LLM-driven framework using prompt tuning and in-context learning, cutting development overhead and accelerating iteration by refining prompts or examples.

Crowdsourcing normalization can be expensive, and it takes weeks of review cycles. PolyNorm, combined with few-shot in-context learning, achieves high grapheme-to-phoneme accuracy with far fewer labeled examples, lowering the barrier for TN in low-resource languages. Beyond cost and efficiency gains, LLM-based TN offers more consistent outputs, better generalization, and language-agnostic deployment, while allowing localized examples to boost performance. By reducing manual effort and speeding iteration, PolyNorm lets developers focus on quality and localization, accelerating research and production.

\subsection{Future Work}
\begin{CJK*}{UTF8}{min}
Future work for PolyNorm includes enabling diacritization restoration in languages like Arabic and Hebrew, where diacritics serve critical grammatical and phonetic roles, and missing markers affect grammar and pronunciation. Targeted prompting could improve accuracy and naturalness of speech synthesis. Japanese and other non-whitespace languages could benefit from integrated tokenizers and multitask learning. While normalizing input to Katakana resolves homograph ambiguities (e.g., 〇 read as "rei," "zero," or "maru") in Japanese, it may disrupt pitch accent patterns crucial for naturalness. Future work should incorporate suprasegmental features like pitch accent and tone prediction into the normalization pipeline.
\end{CJK*}

\section{Conclusion}

We present PolyNorm, a multilingual, LLM-assisted text normalization system designed to address scalability, cost, and language-coverage limitations of  rule-based systems. It achieves strong normalization performance across languages and categories, and significantly reduces WER compared to production-grade rule-based systems, demonstrating its robustness and adaptability.

Beyond performance, PolyNorm represents a paradigm shift in how normalization systems can be developed, maintained, and scaled. The use of instruct prompting and in-context learning enables rapid iteration cycles, eliminates the dependency on handcrafted rules, and lowers the entry barrier for low-resource languages and domain-specific applications. Our language-agnostic prompt design and PolyNorm-Benchmark further establish a foundation for standardized evaluation and future research in LLM-based normalization. As we look ahead, expanding our system to handle complex linguistic phenomena such as diacritization, non-whitespace tokenization, and suprasegmental features will further broaden its applicability, making high-quality TTS systems accessible across a wider range of languages and user contexts.

\section{Acknowledgement}
We are grateful to Alistair Conkie and Xu Shao for their thorough review and valuable feedback that significantly improved this research.

\section{Ethical Consideration}
\subsection{Lay Summary}
This project explores using large language models (LLMs) to automate text normalization—a key step in processing raw text such as speech transcripts, handwriting, or informal writing. Text normalization standardizes text by handling tasks from formatting dates and numbers to restoring diacritics in accented languages. Our LLM-based system offers a scalable, efficient alternative by combining raw input with tailored prompts and few-shot examples. This approach improves consistency and quality across multiple languages, making it well-suited for applications like text-to-speech, machine translation, and speech recognition.

\subsection{AI Safety Disclosure}
LLM-based normalization introduces ethical and safety risks. Because the model learns from large datasets, it may replicate biases in the training data. For example, in gendered languages like Arabic, ambiguous inputs such as ``ant'' (which could mean ``you [masc.]'' or ``you [fem.]'') might be resolved in biased ways. If unaddressed, such biases can reinforce stereotypes or generate inappropriate outputs. Inaccurate normalization may also mislead learners, reinforce incorrect usage, and reduce trust in downstream systems. Therefore, while PolyNorm offers efficient, multilingual normalization, it must be developed with careful attention to data quality, bias mitigation, and user safety to ensure ethical deployment.

\section{Limitations}
While PolyNorm demonstrates strong potential in terms of efficiency and multilingual consistency, the model’s performance is highly dependent on the quality and representativeness of data used as in-context learning examples. Categories covered in the examples may not capture the full range of normalization needs across different languages and varieties. Furthermore, errors in normalization can propagate into downstream tasks such as translation, sentiment analysis, or educational tools. Although expert-reviewed examples were used for in-context learning, the model may still struggle with edge cases or uncommon linguistic patterns. These limitations must be addressed before considering the system for widespread or critical work.

\bibliographystyle{acl_natbib}
\bibliography{custom}

\newpage
\onecolumn
\appendix 
\section*{Appendix}
\section{Instruction Prompt Template}
\label{app:first}

This section presents the instruction prompt template. Note that certain locales include supplementary prompts tailored to their linguistic or stylistic characteristics. In-context learning (ICL) examples are omitted here for brevity.\\ \\
You are an accurate text normalizer for \{locale\}. Your task is to normalize unstandardized text from the following categories to truly reflects how the real speech is, based on the context:\\ 
- Cardinal \\
- Date\\
- Decimal\\
- Ordinal\\
- Fraction\\
- Time\\
- Currency\\
- Unit (Measure)\\
- Electronic Address (URL or Email)\\
- Initialism or Acronym\\
- ISBN\\
- Roman Numeral\\
- Telephone\\
- Sports Score\\
- Mathematical Expression\\
- Symbol\\
- Abbreviation\\
- Chemical Formula\\
- Legal Reference\\
- Vehicle or Product Code\\
- Geographic Coordinates\\
- Version Number\\
- License Plate or Serial Number\\
- Musical Notation\\
- Stock Ticker\\
- Biological Classification\\
- Address\\
- Other unnormalized text\\\\
Some important rules: \\
- When normalizing acronyms, spell out to their full forms for clarity, except when the acronym is a widely recognized and pronounceable name (e.g. “NASA” or “NASCAR”). In those cases, keep the acronym as-is and pronounce it as a word. \\
- If the acronym combines a letter and a word, split accordingly. \\
- Convert punctuation that is spoken aloud into words. For example, write ‘dot’ instead of a period in URLs and emails. \\
- To ensure clarity, segment compound words, websites and file names into recognizable component words rather than keeping them as a whole word. \\
- Symbols in a file name should be read as is. \\
- Common file extensions (.jpeg, .jpg, .txt, etc) should be spoken out. Uncommon file extensions should be spelled out. \\

\section{Baseline Normalization Examples}
\label{app:baseline:examples}

\begin{table*}[h!]

\small
\begin{tabularx}{\textwidth}{p{1.8cm} p{2.2cm} X X X}
\toprule
\textbf{Language} & \textbf{Category} & \textbf{Example} & \textbf{Ground Truth} & \textbf{Error} \\
\midrule
German & Address & Die Wohnung befindet sich in der Hauptstraße 45. & die wohnung befindet sich in der hauptstrasse fünfundvierzig. & die wohnung befindet sich in der hauptstrasse fünf und vierzig. \\
American English & Legal References & The regulation is 15 CFR Part 12. & the regulation is fifteen c f r part twelve. & the regulation is fifteen cfr part twelve. \\
Mexican Spanish & Phone Numbers & Mi celular es 442-789-0123. & mi celular es cuatro cuatro dos, siete ocho nueve, cero uno dos tres. & mi celular es cuatrocientos cuarenta y dos, setecientos ochenta y nueve, cero un veintitrés. \\
French & Sports Scores & Le tennis de table a fini 11-9. & le tennis de table a fini onze à neuf. & le tennis de table a fini onze moins neuf. \\
Italian & Currencies & La tariffa è CHF 20. & la tariffa è venti franchi svizzeri. & la tariffa è c h f venti. \\
Lithuanian & Mathematical Expressions & 5 × 6 = 30. & penki kart šeši lygu trisdešimt. & penki kart šeši yra trisdešimt. \\
Mandarin Chinese & Musical Notation & \begin{CJK*}{UTF8}{gbsn}速度标记\quarternote=120\end{CJK*} & \begin{CJK*}{UTF8}{gbsn}速度标记四分音符等于一百二十\end{CJK*} & \begin{CJK*}{UTF8}{gbsn}速度 标记 = 一 百 二 十\end{CJK*} \\
Japanese & Currencies & \begin{CJK*}{UTF8}{min}価格は\textwon 15,000。\end{CJK*} & \begin{CJK*}{UTF8}{min}カカクワ イチマン ゴセンワン。\end{CJK*} & \begin{CJK*}{UTF8}{min}カカク ハ イチマン ゴセンワン。\end{CJK*} \\
\bottomrule
\end{tabularx}
\caption{Examples of baseline TN errors across the targeted languages and selected categories.}
\label{tab:baseline:tn-errors}
\end{table*}

\section{Baseline model WER (\%) by Language and Selected Categories}
\label{app:baseline:wer}
The Overall column reports the average WER across all 27 categories.

\begin{table*}[h!]
\centering
\small
\begin{tabularx}{\textwidth}{l *{8}{>{\centering\arraybackslash}X}}
\toprule
\textbf{Language} & \textbf{Overall} & \textbf{Address} & \textbf{Legal Ref.} & \textbf{Currencies} & \textbf{Math. Expr.} & \textbf{Musical Not.} & \textbf{Phone Num.} & \textbf{Sports Scores} \\
\midrule
German            & 10.74 & 7.78  & 16.03 & 19.84 & 24.26 & 5.88  & 0.89  & 15.20 \\
American English  & 9.84  & 8.13  & 13.19 & 1.94  & 15.48 & 10.62 & 1.14  & 11.11 \\
Mexican Spanish   & 11.92 & 2.24  & 6.43  & 13.95 & 24.05 & 14.81 & 23.11 & 0.00  \\
French            & 9.72  & 3.19  & 5.75  & 7.14  & 23.95 & 14.63 & 4.98  & 11.88 \\
Italian           & 15.02 & 5.23  & 13.08 & 15.20 & 28.57 & 17.12 & 22.80 & 4.62  \\
Lithuanian        & 10.04 & 9.63  & 12.08 & 7.87  & 15.75 & 18.63 & 16.56 & 6.34  \\
Mandarin Chinese  & 11.36 & 0.00  & 10.65 & 23.48 & 25.00 & 4.23  & 15.72 & 5.92  \\
Japanese          & 17.49 & 16.19 & 19.30 & 18.85 & 20.30 & 18.06 & 15.94 & 21.43 \\
\bottomrule
\end{tabularx}
\caption{Baseline model WER results by language and category}
\label{tab:baseline:results}
\end{table*}

\section{GPT-4o WER (\%) by Language and Selected Categories}
The Overall column reports the average WER across all 27 categories.

\subsection{Iteration 2}
\label{app:gpt:wer2}

\begin{table*}[h!]
\centering
\small
\begin{tabularx}{\textwidth}{l *{8}{>{\centering\arraybackslash}X}}
\toprule
\textbf{Language} & \textbf{Overall} & \textbf{Address} & \textbf{Legal Ref.} & \textbf{Currencies} & \textbf{Math. Expr.} & \textbf{Musical Not.} & \textbf{Phone Num.} & \textbf{Sports Scores} \\
\midrule
German            & 4.24 & 1.11 & 22.90 & 9.52 & 19.12  & 2.52  & 2.23  & 9.31 \\
American English  & 5.23 & 1.44 & 4.68 & 1.02 & 2.38 & 3.54 & 0.38 & 1.85 \\
Mexican Spanish   & 8.45 & 0.45 & 7.86 & 13.95 & 7.59 &  19.75 & 5.78 & 6.25  \\
French            & 5.73 & 0.53 & 4.42 & 1.95 & 11.98 & 6.50 & 2.14 & 1.88 \\
Italian           & 5.12 & 0.58 & 9.23 & 0.80 & 20.17 & 4.50 & 3.20 & 3.08 \\
Lithuanian        & 12.22 & 9.63 & 13.75 & 10.67 & 21.26 & 33.33 & 13.77  & 5.41 \\
Mandarin Chinese  & 6.33 & 0.57 & 7.41 & 18.18 & 10.00 & 4.23 & 11.27 & 5.19 \\
Japanese          & 12.32 & 13.12 & 11.32 & 10.45 & 9.10 & 15.14 & 9.52 & 11.73 \\
\bottomrule
\end{tabularx}
\caption{GPT-4o WER results for Iteration 2}
\label{tab:gpt:iteration2}
\end{table*}

\subsection{Iteration 3}
\label{app:gpt:wer3}

\begin{table*}[h!]
\centering
\small
\begin{tabularx}{\textwidth}{l *{8}{>{\centering\arraybackslash}X}}
\toprule
\textbf{Language} & \textbf{Overall} & \textbf{Address} & \textbf{Legal Ref.} & \textbf{Currencies} & \textbf{Math. Expr.} & \textbf{Musical Not.} & \textbf{Phone Num.} & \textbf{Sports Scores} \\
\midrule
German            & 4.17 & 0.56 & 19.85 & 9.52 & 17.65 & 2.52 & 1.79 & 7.78 \\
American English  & 4.28 & 1.91 & 5.53 & 1.29 & 1.19 & 3.54 & 1.14 & 1.23 \\
Mexican Spanish   & 7.69 & 0.90 & 7.14 & 13.95 & 5.70 & 20.99 & 4.44 & 11.61 \\
French            & 5.65 & 2.66 & 4.42 & 1.30 & 10.78 & 4.07 & 3.20 & 1.88 \\
Italian           & 4.56 & 0.58 & 8.46 & 1.60 & 17.65 & 3.60 & 4.40 &  1.54 \\
Lithuanian        & 6.99 & 5.98 & 12.92 & 7.30 & 17.32 & 16.67 & 8.78 & 4.13 \\
Mandarin Chinese  & 5.05 & 1.06 & 4.51 & 17.76 & 8.29 & 7.04 & 1.59 & 2.59  \\
Japanese          & 7.88 & 3.56 & 9.22 & 6.58 & 7.18 & 8.49 & 4.71 & 8.31 \\
\bottomrule
\end{tabularx}
\caption{GPT-4o WER results for Iteration 3}
\label{tab:gpt:iteration3}
\end{table*}


\end{document}